\documentclass[manuscript,screen]{acmart}

\usepackage[bottom]{footmisc}
\usepackage{url}
\AtBeginDocument{\urlstyle{tt}}
\usepackage{tcolorbox}
\usepackage{float}  % for the H placement option

\newcounter{boxcounter}
\renewcommand{\theboxcounter}{\arabic{boxcounter}}

\newenvironment{floatbox}[1]{%
  \refstepcounter{boxcounter}%
  \begin{tcolorbox}[
    colback=gray!8,
    colframe=gray!50,
    title={\small\textbf{Box~\theboxcounter: #1}},
    fonttitle=\small,
    left=6pt, right=6pt, top=4pt, bottom=4pt,
    boxrule=0.5pt
  ]%
}{%
  \end{tcolorbox}%
}

\acmJournal{TSC}

\begin{document}

\title[Wisdom of LLM Crowds]{Wisdom of LLM Crowds: Aggregation and Contamination in Language Model Ensembles}

\author{Igor Douven}
\email{igor.douven@sorbonne-universite.fr}
\orcid{0000-0003-3413-080X}
\affiliation{%
  \institution{SND\,/\,CNRS\,/\,Sorbonne University}
  \city{Paris}
  \country{France}
}

\begin{abstract}
  The wisdom of crowds---the finding that aggregating judgments across individuals often outperforms the best individual---has been extensively studied with human forecasters. Whether the same phenomenon emerges when the ``crowd'' consists of large language models (LLMs) is an open question with both theoretical and practical implications. We elicited probability estimates from 15~LLMs on 254~binary prediction market questions and evaluated classical and learned aggregation methods. Learned aggregators---a multilayer perceptron and a logistic regression---achieved lower Brier scores than all individual models and classical methods. The logistic regression was found to perform comparably to the neural network, suggesting that the benefit of learned aggregation derives from learning a linear combination of diverse model outputs rather than from nonlinear interactions. Symbolic regression applied to the neural network's learned mapping recovered a pure model-disagreement signal as the lowest-complexity useful formula on the Pareto frontier, further supporting this interpretation. Training cutoff contamination proved a pervasive confound: the apparent capability gap between frontier cloud models and smaller local models collapsed from 35.8\% to 8.9\% on a clean subset of questions resolving after all models' training cutoffs, and individual model rankings showed only moderate stability. Even when the prediction market is evaluated at each model's training cutoff, LLMs remained substantially less accurate, indicating a genuine gap in collective information aggregation. These findings suggest that LLM crowds can exhibit wisdom-of-crowds effects, but that contamination-free evaluation is essential for reliable assessment.
\end{abstract}




\maketitle

%% ============================================================
\section{Introduction}
\label{sec:intro}
%% ============================================================

The wisdom of crowds is the idea that crowds can be smarter than their members \citep{Becker2019}, where this has been understood as meaning that the crowd's aggregate opinion is more accurate (in some sense; see below) than that of its ``average member''  or even that it is more accurate than that of the crowd's \emph{best} members. There is ample evidence for the former claim \citep{Galton1907,Surowiecki2004,DavisStober2014} and some for the latter \citep{HongPage2004,Page2007,Prelec2017,Douven2019,Douven2022,HongPage2025}, and the mechanisms underlying this type of crowd wisdom are also well understood, at least in broad terms: individual forecasters make partially independent errors, and averaging cancels those errors while preserving shared signal \citep{LarrickSoll2006}. That said, the reliability of the phenomenon should not be overstated. Recent large-scale analyses of professional forecasting data suggest that, under simple aggregation rules such as the mean or the median, the crowd's advantage over a randomly selected forecaster is often modest, and that the impression of a quasi-lawlike ``wisdom of crowds'' may partly reflect selective attention to striking successes \citep{ReiaFontanari2021,SiqueiraNetoFontanari2023}. Rather than undermining the research program, however, such critiques underline that how much wisdom a crowd exhibits is an empirical matter that depends on the composition of the crowd, the conditions of elicitation, and---importantly for present purposes---on how the individual judgments are aggregated. Indeed, building on the insight about error cancellation, recent work has identified a number of powerful aggregation strategies that specifically exploit systematic features of the error distribution \citep{DouvenKriegeskorteStinson2026}.

With the current broad availability of large language models (LLMs), it becomes natural to ask whether LLMs can substitute for human forecasters in crowd wisdom paradigms.\footnote{On the more general question of whether LLMs can serve as useful models of human cognition and even as ``synthetic participants,'' see \cite{Argyle_2023,BinzSchulz2023,Hagendorff2023,Hashimoto2026}.} If so, a crowd of LLMs could provide a cheap and scalable alternative to human forecasting panels for a wide range of tasks. Moreover, studying LLM crowds could further illuminate the mechanisms of crowd wisdom itself: if error-pattern diversity---understood throughout this paper as weakly correlated errors across crowd members, a measurable property of the crowd's joint error distribution \citep{DouvenKriegeskorteStinsonYing2026}, rather than mere spread among the estimates (see Sect.~\ref{sec:coef})---drives aggregation benefits, then LLM crowds should also benefit from aggregation to the extent that their outputs are sufficiently diverse---even if the degree of diversity, and hence the magnitude of the benefit, may differ from human crowds, whose members bring a much greater variety of backgrounds and expertise \cite{Landemore2020}. Crucially, if diversity is the key ingredient, then even simple learned aggregators that weight models by their historical agreement patterns may suffice---a prediction our results will confirm.

The closest existing work to ours is Schoenegger et al.'s \cite{Schoenegger2024}, which used a crowd of 12~LLMs to make probabilistic predictions on 31~binary questions from a real-time forecasting tournament, finding that the LLM crowd---aggregated by simple median---was statistically indistinguishable from a human crowd of 925~forecasters. This is an important proof of concept, but it leaves some important questions unanswered. First, Schoenegger and colleagues used only a very simple aggregation method (the median) and did not ask whether more sophisticated aggregators could improve over the best individual model or over the median itself. And second, they offered no account of the \emph{mechanism} underlying any aggregation benefit---whether it reflects error-pattern diversity, performance-weighted combination, or something else. It is also to be noted that, because they collected data in real time on future events, contamination was not an issue for their study; our study, which draws on resolved questions across a longer time window, faces a more serious contamination challenge that requires explicit treatment \citep{carlini2021extracting,roberts2024to,xu2024benchmark,gao2025test}.\footnote{\citet{Halawi2024} and \citet{Zou2022} similarly evaluate individual LLMs or simple ensembles on forecasting benchmarks without addressing the aggregation question systematically. \citet{Halawi2024} show that a retrieval-augmented system can approach human crowd accuracy, but their system uses web search rather than crowd aggregation as the source of improvement.}

The present study aims to address these issues. We elicited probability estimates from 15~LLMs---including both locally deployable small models and frontier cloud models---on 254~binary prediction market questions. We evaluated four classical aggregation methods alongside a logistic regression baseline and a trained multilayer perceptron (MLP) aggregator, and applied symbolic regression to recover an interpretable account of the MLP's learned strategy. We also conducted a systematic contamination analysis, examining whether training cutoff overlap inflated apparent individual model performance and distorted model rankings.

\section{Methods}

\subsection{Materials}\label{subsec:mat}

We collected binary (YES/NO) prediction market questions from Manifold Markets (\url{https://manifold.markets}), a play-money forecasting platform that hosts a wide range of questions on current events, science, technology, and politics.\footnote{One might worry that the fact that Manifold is a play-money platform could affect market efficiency relative to real-money markets. However, \citet{Wolfers2004} show that play-money prediction markets produce well-calibrated probabilities comparable to real-money markets. We explored other platforms (Kalshi, Polymarket, Metaculus) as well, but none yielded a sufficient number of questions meeting our selection criteria within the required resolution window.} Questions were selected to have resolved between June~2024 and March~2026 and to have attracted at least 75~unique traders, a threshold chosen to ensure that the question had received sufficient community attention. After converting question titles to declarative statements,\footnote{So, for example, ``Will the US government shut down on Monday?''~became ``The US government will shut down on Monday.''} a subset of questions was excluded on grounds of ambiguity, personal reference, or dependence on inside information not accessible to a language model (e.g., ``Resolves to the side with the most holders at market close''; ``I smoke weed before January 1, 2026''; ``Does cooking in a cast-iron pan make steak better? [Eloise's Blind Taste Test]''). The initial dataset comprised 208~statements, of which 107~resolved true (YES) and 101~resolved false (NO)\@. The mean community probability at market close was 0.53 ($SD = 0.30$).

Because models may have been trained on data that includes the outcomes of questions resolving before their training cutoff (the most recent cutoff among our 15~models was August~2025; see Table~\ref{tab:models}), we defined a \emph{clean subset} consisting of questions resolving after 1~September~2025. Of the 208~items in the main dataset, 48~resolved after this date. To increase statistical power, we supplemented these with 46~additional Manifold Markets questions collected using the same selection criteria and also resolving after 1~September~2025, yielding a clean subset of 94~items (44~YES, 50~NO) used for the primary aggregation analysis and all other contamination-free analyses, and bringing the total number of unique items across both datasets to 254. The remaining 160~items from the initial 208-item dataset (resolving before September~2025) serve as the training set for learned aggregators evaluated on the clean subset (see Sect.~\ref{sec:res}).

\subsection{Language models}

We elicited probability estimates from 15~language models spanning a range of architectures, sizes, and training regimes. Five models were run locally via Ollama; three were accessed via the Anthropic API; three via the OpenAI API; three via the Google AI API; and one via the DeepSeek API\@. See Table~\ref{tab:models} for further details.\footnote{Exact model identifiers used in API calls are available in the elicitation script that is included in the Supplementary Materials.} Because we also wanted to analyze how model capability affects aggregation performance, we included both frontier cloud models and smaller locally deployable models.

\begin{table}[t!!!]
\centering
\caption{Language models included in the study. Training cutoffs are approximate and drawn from official documentation where available. Model sizes for cloud-hosted models are not publicly disclosed.}
\label{tab:models}
\small
\begin{tabular}{l @{\hspace{10mm}} l @{\hspace{10mm}} l @{\hspace{10mm}} l}
\toprule
Model & Provider & Approx.\ size & Training cutoff \\
\midrule
Mistral~7B        & Ollama/Mistral AI  & 7B  & March 2023 \\
LLaMA~3.1~8B      & Ollama/Meta        & 8B  & December 2023 \\
Gemma~2~9B        & Ollama/Google      & 9B  & June 2024 \\
Phi-4~14B         & Ollama/Microsoft   & 14B & June 2024 \\
Qwen~2.5~7B       & Ollama/Alibaba     & 7B  & September 2024 \\
GPT-5.4~Mini      & OpenAI             & --- & September 2024 \\
GPT-5.2           & OpenAI             & --- & September 2024 \\
GPT-5.4           & OpenAI             & --- & September 2024 \\
DeepSeek-chat     & DeepSeek           & --- & November 2024 \\
Gemini Flash-Lite & Google             & --- & January 2025 \\
Gemini Flash      & Google             & --- & January 2025 \\
Gemini Pro        & Google             & --- & January 2025 \\
Claude Haiku~4.5  & Anthropic          & --- & July 2025 \\
Claude Sonnet~4.6 & Anthropic          & --- & August 2025 \\
Claude Opus~4.6   & Anthropic          & --- & August 2025 \\
\bottomrule
\end{tabular}
\end{table}

\subsection{Elicitation procedure}\label{sec:elicitation}

Each model was presented with a standardized prompt asking it to estimate the probability that a given statement is true, to think
briefly, and to conclude with a single number between 0 and~1 in the format \texttt{PROBABILITY: [number]}; see Box~\ref{box:prompt} for
an illustration. All models were queried via their respective APIs (or via Ollama for local models) with \texttt{max\_tokens} set to
512 and temperature left at the provider default (equivalent to~1.0 in all cases). Most models followed the prompt format reliably.\footnote{Google Gemini models initially failed to produce a parseable probability on a substantial fraction of items (Gemini Flash: 88\%; Gemini Pro: 66\%) due to free-form response formatting; a single follow-up turn requesting the probability in the required format recovered most of these. LLaMA~3.1~8B initially refused politically sensitive questions on 22.6\% of items; a brief neutral system message framing the task as an academic calibration exercise recovered most refusals. Mistral~7B produced responses without a \texttt{PROBABILITY:} tag on 46\% of items, but a valid decimal number was ultimately recovered in every case through the retry and fallback procedures.} After all retry and fallback procedures, 12 of the $15\times208 = 3{,}120$ model--item queries on the main dataset (0.4\%) still yielded no usable probability (LLaMA~3.1~8B: 2; Gemma~2~9B: 1; Claude Opus~4.6: 3; Gemini Flash-Lite: 1; Gemini Pro: 5), as did 4 of the $15\times94=1{,}410$ queries on the clean subset. These cells are treated as missing throughout: individual-model statistics (including the Brier scores in Tables~\ref{tab:individual} and~\ref{tab:cutoff} and the within-/outside-cutoff counts in Table~\ref{tab:contamination}, which for this reason do not always sum to 208) are computed over each model's non-missing items, and the classical aggregators are computed row-wise over the model outputs available for a given item. The two learned aggregators (Sect.~\ref{sec:aggmeth}), whose input is the full 15-dimensional vector of model estimates, require complete inputs; for them, missing cells are imputed with the corresponding model's mean estimate computed on the training set, and the same training-set means are used to impute the few missing cells in the evaluation set, so that no preprocessing of the evaluation data depends on the evaluation set itself. (The symbolic regression analysis handles missingness differently; see Sect.~\ref{sec:srmeth}.)

\begin{figure}[b!!!!!]
\begin{floatbox}{Example elicitation prompt}\label{box:prompt}
\small
\texttt{You are being asked to assess the probability that the following
statement is true. The statement was evaluated as of 2025-11-14. Use
this date to interpret any relative time references.}

\medskip
\texttt{STATEMENT: "The Doomsday Clock will be 60s (or less) to midnight
by end of January 2026."}

\medskip
\texttt{Instructions:}\\
\texttt{- This is a binary claim (either true or false).}\\
\texttt{- Give your probability that the statement is TRUE, as a number
between 0 and 1.}\\
\texttt{- Base your answer only on your training knowledge. Do not search
the internet.}\\
\texttt{- Think briefly, then state your final answer on a new line in
the exact format:}\\
\texttt{\phantom{--}PROBABILITY: [number]}\\
\texttt{For example: PROBABILITY: 0.73}

\medskip
\texttt{Your response:}
\end{floatbox}
\end{figure}

\subsection{Aggregation methods}\label{sec:aggmeth}

As argued in \citet{DouvenKriegeskorteStinson2026}, how much wisdom there is in a crowd depends, to a large extent, on how we aggregate the opinions of the individuals constituting the crowd. Following these authors, we evaluated a number of classical aggregation methods as well as a trained neural network aggregator. We also looked at a logistic regression baseline, and we applied symbolic regression to the neural network's learned strategy.

\subsubsection*{Classical aggregators}

Of the four classical methods we evaluated, three are instances of the general $f$-mean family \citep{Bullen2003}, which maps a vector of probabilities $(p_1,\ldots,p_N)$ to a scalar aggregate via
\[
  M_f(p_1,\ldots,p_N) \:\:=\:\: f^{-1}\Bigl(\tfrac{1}{N}
  \textstyle\sum_{i=1}^{N} f(p_i)\Bigr).
\]
Choosing $f$ as the identity gives the \emph{arithmetic mean}, arguably the most natural baseline, and choosing $f(x) = x^{-1}$ gives the \emph{harmonic mean}. Setting $f = \mathrm{logit}$ yields the \emph{log-odds mean},
\[
  \sigma\Bigl(\tfrac{1}{N}\textstyle\sum_i \mathrm{logit}(p_i)\Bigr),
\]
where $\sigma$ denotes the logistic function. The log-odds mean has been advocated as the normative standard for probabilistic aggregation under the assumptions that individual judgments are conditionally independent and well-calibrated \citep{Morris1983, GenestZidek1986,DouvenKriegeskorteStinson2026}, and it corresponds to what is sometimes called \emph{extremizing} the arithmetic mean (i.e., pushing the aggregate away from 0.5 relative to simple averaging). It is worth noting that, for binary outcomes, the log-odds mean is algebraically identical to the geometric mean of the reported probabilities renormalized to sum to one across the two outcomes, given that
\[
  \frac{\bigl(\prod_i p_i\bigr)^{1/N}}
       {\bigl(\prod_i p_i\bigr)^{1/N} + \bigl(\prod_i (1-p_i)\bigr)^{1/N}}
  \:\:=\:\:
  \sigma\Bigl(\tfrac{1}{N}\textstyle\sum_i \mathrm{logit}(p_i)\Bigr),
\]
so we treat this as a single method and refer to it as the \emph{geometric/log-odds mean}. Finally, we include the \emph{median}, which was the primary aggregator in \citet{Schoenegger2024} and is more robust to outlier estimates than the arithmetic mean.

\subsubsection*{Neural network aggregator (MLP)}

We followed \citet{DouvenKriegeskorteStinson2026} in going beyond the closed-form classical aggregators and training a multilayer perceptron (MLP) to learn an aggregation function directly from the data. The network maps the 15-dimensional vector of model probability estimates to a single aggregate probability via two hidden layers: input (15~units) $\rightarrow$ hidden (64~units, ReLU activation, dropout $p = 0.2$) $\rightarrow$ hidden (32~units, ReLU, dropout $p = 0.2$) $\rightarrow$ sigmoid output. The network was trained over 25 epochs using the Adam optimizer with learning rate $5\times10^{-3}$, binary cross-entropy loss, and batch size~16. For the aggregation analysis on the clean subset (Sect.~\ref{subsec:mat}), the MLP is trained on the 160~non-clean items and evaluated on all 94~clean items (temporal transfer). The relatively simple architecture was chosen after preliminary experiments showed no benefit from larger networks, consistent with the limited training data available.

\subsubsection*{Logistic regression baseline}

To determine whether the MLP's advantage over classical means requires nonlinear processing, we also trained an L2-regularized logistic regression ($\lambda = 0.01$) on the same 15-dimensional input vector. Logistic regression is a linear learned aggregator: it can learn non-uniform weightings of the models but cannot combine them nonlinearly. Like the MLP, it was trained on the 160~non-clean items and evaluated on all 94~clean items in temporal transfer mode, ensuring a fair comparison. If logistic regression performs comparably to the MLP, that is evidence that the aggregation benefit is attributable to learning an appropriate linear combination of diverse model outputs, rather than to deep nonlinear feature interactions.

One aspect of this training design deserves to be mentioned at the outset. The 160~training items all resolve before September~2025 and are therefore precisely the items subject to potential training-cutoff contamination (Sect.~\ref{sec:contmeth}). Because contamination alters model error patterns, this creates distribution shift: weights fitted on potentially contaminated historical items are applied to genuinely novel questions. Note that this is, if anything, a handicap the learned aggregators must overcome rather than an advantage, and that contamination-free \emph{evaluation}---which both learned aggregators receive, being evaluated exclusively on the clean subset---is what matters for the paper's argument. We come back to this point in Section~\ref{sec:limits}.

\subsection{Symbolic regression}\label{sec:srmeth}

To obtain a still better understanding of the MLP's learned aggregation strategy---and, in particular, to obtain evidence on the question whether learned aggregation amounts to covert expert selection, a question we take up in Sections~\ref{sec:srres} and~\ref{sec:woc_or_expert}---we took our cue from \citet{Cranmer2020} and applied symbolic regression to the trained network. Symbolic regression is a machine-learning technique that searches the space of mathematical expressions for a formula that best approximates a target function \citep{Koza1992,Schmidt2009}. Rather than fitting parameters within a fixed functional form, it simultaneously searches over the form and the parameters, representing candidate formulas as trees whose leaves are input variables or constants and whose internal nodes are mathematical operations (addition, multiplication, logistic function, and so on). A population of candidate formulas is evolved over many generations via selection, mutation, and crossover, gradually improving both accuracy and simplicity. (See \citet{Cranmer2020} for details.)

When, as in our case, simplicity and accuracy trade off against each other, the search naturally produces a \emph{Pareto frontier}---a set of formulas such that no formula achieves better accuracy without becoming more complex, and no formula is simpler without sacrificing accuracy. For this analysis, we retrained the MLP on all complete rows of the full 208-item dataset ($n = 196$; here, no imputation is applied, so the 12~rows containing a missing cell are dropped) rather than on the 160~non-clean items used for the temporal transfer evaluation. This is appropriate because the goal here is not out-of-sample predictive performance but \emph{interpretability}: training on the maximum available data gives the MLP a more stable learned mapping for SR to approximate. Contamination may of course shape the mapping learned from the full dataset (cf.\ Sect.~\ref{sec:limits}); our purpose here is simply to characterize that mapping, not to claim that it is contamination-free. We then used the \texttt{SymbolicRegression.jl} package \citep{Cranmer2023} to search for a symbolic formula minimizing mean squared error against this MLP's predictions on those same 196~items, with formula complexity penalized alongside predictive error. Given that SR is a stochastic process, we conducted 50~independent SR runs and examined the Pareto frontier of each to assess which formulas appeared consistently in the runs.

Two clarifications about the logic of this analysis are in order. First, the SR target is, by design, the MLP's predictions rather than the ground-truth outcomes: SR is used here as a model-distillation tool in the sense of \citet{Cranmer2020}, whose purpose is to characterize the function the MLP has learned, not to construct an independent predictive model. Ground truth enters at the evaluation stage instead: any formula recovered by SR is subsequently scored against the actual resolutions, on both the full dataset and the clean subset (Sect.~\ref{sec:srres}), so no performance claim in this paper rests on a formula's agreement with the MLP\@. Second, because SR is stochastic and the space of candidate formulas is vast, reporting the output of a single run would invite concerns about post-hoc selection; the 50~independent runs, and the selection frequencies computed across them (Table~\ref{tab:sr}), are meant to preempt this concern by showing which components of the recovered formulas are stable and not run-specific.

\subsection{Evaluation}

Our primary evaluation metric is the Brier score,
\begin{equation*}
  \text{BS} \:\: = \:\: \frac{1}{N}\sum_{i=1}^{N} (p_i - y_i)^2,
\end{equation*}
where, for each statement~$i$ ($1\leqslant i\leqslant N$), $p_i \in [0,1]$ is the probability assigned to it and $y_i \in \{0,1\}$ its truth value. While the word ``score'' might suggest otherwise, BS is actually a \emph{penalty}, meaning that lower is better \citep{Brier1950}. As secondary metrics we report classification accuracy---the proportion of items on which the model's prediction exceeds 0.5 precisely if the statement resolved true---and the familiar area under the receiver operating characteristic curve (AUC), which measures the model's ability to rank true statements above false ones regardless of the absolute probability values. The three metrics provide complementary perspectives: Brier score rewards both calibration and resolution, accuracy rewards threshold-level discrimination, and AUC rewards rank-order discrimination \citep{gneiting2007,DOUVENKRIEGESKORTESCORING}.

\subsection{Contamination analysis}\label{sec:contmeth}

As mentioned previously, in evaluating LLMs on resolved prediction market questions there is the concern of training cutoff contamination; specifically, that if a question's outcome was publicly reported before a model's training cutoff, the model may have seen that outcome during training and so, when prompted, may simply provide the ``memorized'' answer rather than reason about the prompt \citep{Deng2024,roberts2024to}. This could inflate apparent individual model performance and distort comparisons between models with different cutoff dates; most notably, it could favor newer frontier models over older or smaller ones for reasons unrelated to reasoning ability.

To assess the extent and consequences of contamination in our dataset, we conducted three analyses. First, for each model we compared the probability extremity ($|p - 0.5|$) on within-cutoff items (questions resolving before the model's training cutoff) versus outside-cutoff items, using Welch $t$-tests; elevated extremity on within-cutoff items would indicate overconfident retrieval of memorized outcomes. Second, we compared classification accuracy within versus outside each model's training window; a contamination effect should manifest as substantially higher accuracy on within-cutoff items. Finally, we computed Spearman rank correlations between model rankings on the full 208-item dataset and on the 94-item clean subset (Sect.~\ref{subsec:mat}) to quantify how much contamination distorts the apparent ordering of models.

\subsection{Comparison with the Manifold crowd}\label{sec:manifold}

\citet{Schoenegger2024} compared their LLM crowd directly against a human forecasting panel, finding the two statistically indistinguishable. Our design differs in an important respect: rather than a forecasting tournament, we use a continuously updated prediction market, which means a naive comparison against the final market probability would be unfair to the LLMs---the market incorporates information accumulated right up to resolution, while LLMs reason from a frozen training snapshot. We therefore construct two baselines. The \emph{final} market probability is the closing price at resolution, and serves as an upper bound on human crowd performance. The \emph{cutoff-matched} market probability is the last recorded price on or before midnight UTC of each model's training cutoff date, providing a temporally matched comparison wherever a pre-cutoff market price is available; the exceptional fallback cases are discussed below. Since training cutoffs are reported at month granularity (Table~\ref{tab:models}), we operationalize each cutoff throughout---here and in the contamination analyses of Section~\ref{sec:contmeth}---as the first day of the reported month.

Cutoff-matched probabilities were retrieved via the Manifold API's bet history endpoint, which returns the full sequence of bets with timestamps. For each market and each model, we took the market probability immediately after the last bet placed on or before the cutoff date (i.e., the running consensus price at that point in time, reflecting the aggregate of all trading activity up to the cutoff). When a market had received no bets before a model's training cutoff---either because the market was created after that date or because it had not yet attracted any traders---we fell back to the market's initial probability (the \texttt{initialProbability} field from the market endpoint), which on Manifold is set by the market creator at opening and defaults to~0.5. This fallback occurred for approximately 55 markets (26\% of the dataset) for the four models with the earliest cutoffs (Mistral~7B, LLaMA~3.1~8B, Gemma~2~9B, and Phi-4~14B, all with cutoffs in 2023 or June~2024); for models with later cutoffs the fallback rate was substantially lower, though it did not vanish entirely: approximately 13~items (6.3\%) still required the fallback at the latest (August~2025) cutoff. All 208~items were successfully covered for all 15~models, with no missing cutoff-matched probabilities.\footnote{A caveat applies to the fallback cases: when a market was created only \emph{after} a model's cutoff, its initial probability was necessarily set after that date, so the corresponding market--model comparison is not strictly cutoff-matched and may advantage the market, since the creator's opening price can reflect information unavailable at the cutoff. As a sensitivity check, we recomputed the ratios of Table~\ref{tab:cutoff} excluding, for each model, all items whose market probability shows no movement across any cutoff date up to and including that model's (for Mistral~7B, whose cutoff is the earliest, we applied the December~2023 criterion). Because a market with no bets before a given cutoff necessarily shows a constant price across all earlier cutoffs, this excludes a superset of the fallback cases. The market's advantage persists for every model, with ratios ranging from 1.40 to 2.68 (mean: 1.94; versus 1.61--2.64 and 1.95 on all items); as expected, the largest reductions occur for the earliest-cutoff models (e.g., Mistral~7B: $1.80\times \to 1.50\times$). The arithmetic-mean comparison at the August~2025 cutoff is essentially unaffected: only 13~items (6.3\%) are excluded by this conservative criterion at that date, and excluding them changes the ratio from $2.01\times$ to $2.03\times$.\label{fn:fallback}}

%------------------------------------------------------------
\section{Results}\label{sec:res}

For completeness and comparability with prior work, Table~\ref{tab:individual} reports model performance for all 15~models on the full 208-item dataset. We see that Claude Sonnet~4.6 achieved the lowest Brier score among individual models (BS = 0.273), followed by Claude Opus~4.6 (BS = 0.284) and Gemini Flash (BS = 0.286). Local models showed substantially worse performance (Brier scores 0.362--0.413).

\subsection{Aggregation performance}\label{sec:aggres}

\begin{table}[t!!!]
\centering
\caption{Individual model performance on the initial 208-item dataset. Models per category (cloud/local) sorted by Brier score.}
\label{tab:individual}
\small
\begin{tabular}{l @{\hspace{11mm}} l @{\hspace{11mm}} l
                @{\hspace{11mm}} r @{\hspace{11mm}} r}
\toprule
Model & Group & Brier & Accuracy & AUC \\
\midrule
Claude Sonnet~4.6   & Cloud & 0.273 & 0.635 & 0.677 \\
Claude Opus~4.6     & Cloud & 0.284 & 0.615 & 0.668 \\
Gemini Flash        & Cloud & 0.286 & 0.630 & 0.640 \\
Gemini Pro          & Cloud & 0.292 & 0.635 & 0.693 \\
Claude Haiku~4.5    & Cloud & 0.311 & 0.524 & 0.532 \\
GPT-5.4~Mini        & Cloud & 0.328 & 0.587 & 0.632 \\
GPT-5.2             & Cloud & 0.338 & 0.548 & 0.534 \\
GPT-5.4             & Cloud & 0.347 & 0.553 & 0.538 \\
DeepSeek-chat       & Cloud & 0.361 & 0.524 & 0.515 \\
Gemini Flash-Lite   & Cloud & 0.377 & 0.580 & 0.584 \\
\midrule
Gemma~2~9B          & Local & 0.362 & 0.502 & 0.450 \\
Phi-4~14B           & Local & 0.381 & 0.447 & 0.391 \\
Qwen~2.5~7B         & Local & 0.389 & 0.418 & 0.431 \\
LLaMA~3.1~8B        & Local & 0.394 & 0.427 & 0.430 \\
Mistral~7B          & Local & 0.413 & 0.469 & 0.452 \\
\bottomrule
\end{tabular}
\end{table}

Naturally, our main current interest is not so much in the individual models but rather in aggregation performance. For the aggregation analysis, we turn to the 94-item clean subset, on which all methods are evaluated consistently. The classical aggregators require no training and are evaluated directly on all 94~items. The two learned aggregators---logistic regression and MLP---are trained and evaluated in the temporal transfer design described in Section~\ref{sec:aggmeth}, so they are applied to questions they have never seen. This design mirrors real-world deployment---an aggregator fitted on resolved historical questions is applied to new ones---and ensures that all methods are compared on identical ground.

\begin{figure}[t!!!!!]
\centering
\includegraphics[width=.775\textwidth]{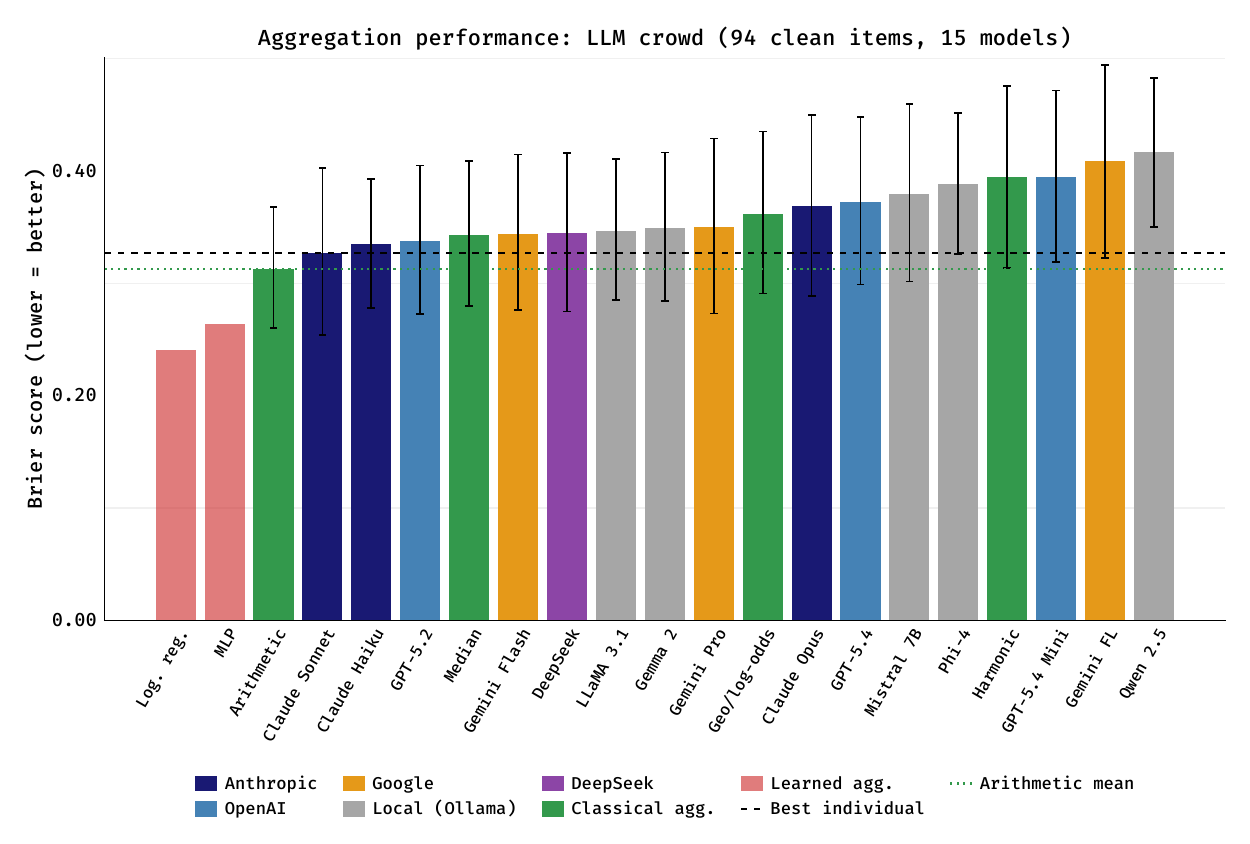}
\caption{Brier scores on the 94-item clean subset for the 15~models and the aggregators (sorted by Brier score). Both learned aggregator scores are from temporal transfer (trained on 160 non-clean items, tested on all 94~clean items), making them comparable to the all-data estimates for classical aggregators. Error brackets show 95\% bootstrap confidence intervals for individual models and classical aggregators. No brackets are drawn for the learned aggregators since their temporal transfer evaluation design is not comparable to the all-data bootstrap used for other methods; their statistical advantage is instead assessed via the sign and permutation tests reported in the text.}
\label{fig:perf}
\end{figure}

The two learned aggregators showed complementary strengths: logistic regression achieved a lower Brier score (BS = 0.241 vs.~BS = 0.264 for the MLP), while the MLP showed better rank-order discrimination (AUC = 0.657 vs.~0.633) and threshold-level accuracy (0.606 vs.~0.574). Both substantially outperformed the arithmetic mean on all three metrics (Brier: 0.313; accuracy: 0.511; AUC: 0.498). The near-equivalence of the two learned aggregators across metrics is consistent with the aggregation benefit being largely captured by a linear combination of model outputs, with any residual nonlinearity in the MLP improving discrimination at the cost of calibration. The MLP's advantage over the arithmetic mean was statistically significant across items (sign test: $62/94$, $p = .002$; Wilcoxon: $z = -2.48$, $p = .013$; permutation test: $p = .049$). The logistic regression's larger mean Brier reduction (23\% vs.~15.6\% for the MLP) was not item-consistently significant (sign test: $46/94$, $p = .837$; Wilcoxon: $z = -1.55$, $p = .122$), though the permutation test, which is sensitive to the mean reduction rather than item-level consistency, \emph{was} significant ($p = .002$). This pattern suggests that logistic regression's advantage is concentrated on a smaller number of items on which it wins decisively, rather than distributed uniformly across the clean subset. Figure~\ref{fig:perf} shows Brier scores for all methods.

\begin{table}[t!!!!]
\centering
\caption{Crowd-split analysis: Brier scores for the local-only crowd (5~models) and the cloud-only crowd (10~models) under each aggregation method, on the full 208-item dataset and on the 94-item clean subset. Diff is the percentage by which the local crowd's Brier score exceeds the cloud crowd's, that is, $100\,(\text{BS}_{\text{local}} - \text{BS}_{\text{cloud}})/\text{BS}_{\text{cloud}}$. The arithmetic-mean entries are the cloud--local gaps referred to in the text (35.8\% on the full dataset, 8.9\% on the clean subset).}
\label{tab:crowdsplit}
\small
\begin{tabular}{l @{\hspace{8mm}} r @{\hspace{6mm}} r @{\hspace{6mm}} r @{\hspace{10mm}} r @{\hspace{6mm}} r @{\hspace{6mm}} r}
\toprule
& \multicolumn{3}{c@{\hspace{10mm}}}{Full dataset (208 items)} & \multicolumn{3}{c}{Clean subset (94 items)} \\
\cmidrule(r{9mm}){2-4}\cmidrule{5-7}
Method & Local & Cloud & Diff (\%) & Local & Cloud & Diff (\%) \\
\midrule
Arithmetic mean          & 0.345 & 0.254 & +35.8 & 0.340 & 0.312 & +8.9 \\
Median                   & 0.368 & 0.275 & +33.6 & 0.354 & 0.343 & +3.1 \\
Geometric/log-odds mean  & 0.399 & 0.287 & +39.3 & 0.378 & 0.360 & +4.9 \\
Harmonic mean            & 0.407 & 0.366 & +11.1 & 0.368 & 0.378 & $-$2.7 \\
\bottomrule
\end{tabular}
\end{table}

To examine whether the capability gap between cloud and local models persists under aggregation, we computed Brier scores separately for the cloud-only crowd (10~models) and the local-only crowd (5~models) under each classical aggregator, on both the full 208-item dataset and the 94-item clean subset; see Table~\ref{tab:crowdsplit}. On the full dataset, the cloud crowd outperformed the local crowd by a wide margin under most methods (e.g., 35.8\% under arithmetic mean aggregation: BS = 0.254 vs.~0.345). On the clean subset, cloud models outperformed local models by only 8.9\% under arithmetic mean aggregation (BS = 0.312 vs.~0.340), a gap that largely disappears for the other classical methods; we return to this contrast in Section~\ref{sec:contres}. The median---the aggregator used by \citet{Schoenegger2024}---achieved a Brier score of 0.343, outperforming the geometric/log-odds and harmonic means but falling below the arithmetic mean (0.313). This suggests that simple averaging captures more signal than the median in our setting, where model outputs vary continuously rather than clustering at extreme values.

\subsection{Training cutoff contamination}\label{sec:contres}

\begin{table}[t!!!]
\centering
\caption{Contamination check: model accuracy on within-cutoff items (resolving before the model's training cutoff) versus outside-cutoff items. Models with zero within-cutoff items in the dataset are omitted. Asterisks mark models showing significantly greater extremity, $|p - 0.5|$, on within-cutoff items according to Welch $t$-tests ($p < .05$). For models with missing (unparseable) responses (Sect.~\ref{sec:elicitation}), $n_{\text{in}} + n_{\text{out}}$ falls short of 208 by the number of missing items (Gemini Flash-Lite: 1; Gemini Pro: 5; Claude Opus~4.6: 3).}
\label{tab:contamination}
\small
\begin{tabular}{l @{\hspace{10mm}} r @{\hspace{10mm}} r @{\hspace{10mm}} r @{\hspace{10mm}} r @{\hspace{10mm}} r @{\hspace{10mm}} r}
\toprule
Model & Cutoff & $n_{\text{in}}$ & Acc$_{\text{in}}$ &
       $n_{\text{out}}$ & Acc$_{\text{out}}$ & $\Delta$Acc \\
\midrule
Qwen~2.5~7B       & Sep 2024 & 27  & 0.556 & 181 & 0.398 & $+0.158$ \\
GPT-5.4~Mini      & Sep 2024 & 27  & 0.630 & 181 & 0.580 & $+0.050$ \\
GPT-5.2           & Sep 2024 & 27  & 0.778 & 181 & 0.514 & $+0.264$ \\
GPT-5.4           & Sep 2024 & 27  & 0.741 & 181 & 0.525 & $+0.216$ \\
DeepSeek-chat     & Nov 2024 & 44  & 0.568 & 164 & 0.512 & $+0.056$ \\
Gemini Flash-Lite & Jan 2025 & 91  & 0.637 & 116 & 0.535 & $+0.103$ \\
Gemini Flash      & Jan 2025 & 92  & 0.728 & 116 & 0.552 & $+0.177$\rlap{$^{*}$} \\
Gemini Pro        & Jan 2025 & 91  & 0.747 & 112 & 0.545 & $+0.203$\rlap{$^{*}$} \\
Claude Haiku~4.5  & Jul 2025 & 153 & 0.575 & 55  & 0.382 & $+0.193$ \\
Claude Sonnet~4.6 & Aug 2025 & 156 & 0.699 & 52  & 0.442 & $+0.256$ \\
Claude Opus~4.6   & Aug 2025 & 155 & 0.652 & 50  & 0.500 & $+0.152$ \\
\bottomrule
\multicolumn{7}{l}{\footnotesize $^{*}$ Significantly elevated extremity
  on within-cutoff items ($p < .05$).}
\end{tabular}
\end{table}

All models with sufficient within-cutoff items showed substantially higher accuracy on those items than on out-of-window items (Table~\ref{tab:contamination}), with accuracy differentials ranging from $+0.05$ (GPT-5.4~Mini) to $+0.26$ (Claude Sonnet). Two models showed significantly elevated probability extremity on within-cutoff items: Gemini Flash ($t = 4.25$, $p < .001$) and Gemini Pro ($t = 2.68$, $p = .008$).

\begin{figure}[b!!!]
\centering
\includegraphics[width=\textwidth]{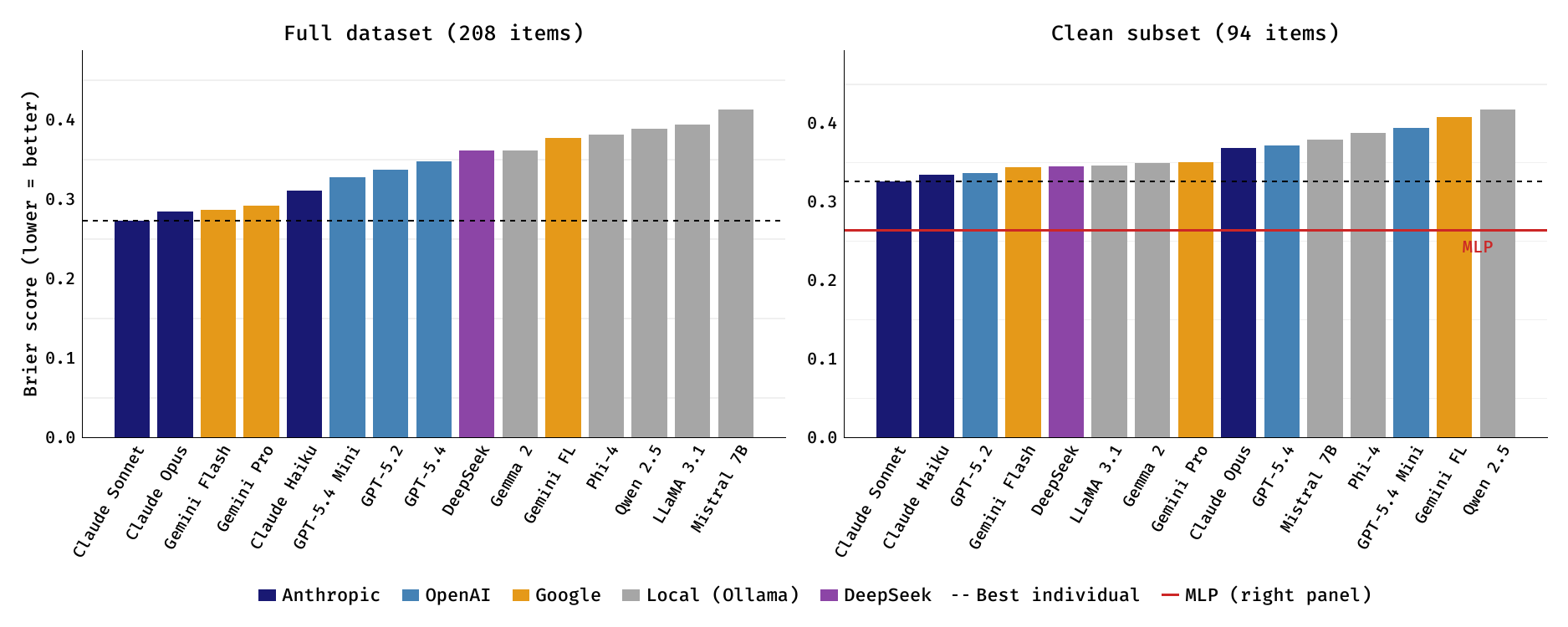}
\caption{Individual model Brier scores on the initial 208-item dataset (left) and the 94-item clean subset (right), sorted by Brier score within each panel. Bars are colored by model family. The dashed line marks the best individual model in each panel; the solid red line in the right panel marks the MLP's temporal-transfer score (BS = 0.264), which is evaluated on the same 94~clean items as the individual models shown there. No corresponding line is drawn in the left panel, as there is no comparable out-of-sample MLP evaluation on the full dataset. Notable rank changes between panels (full~$\rightarrow$~clean): LLaMA~3.1~8B: $14\rightarrow 6$; GPT-5.4~Mini: $6\rightarrow 13$; Claude Opus: $2\rightarrow 9$; GPT-5.2: $7\rightarrow 3$; DeepSeek: $9\rightarrow 5$.}
\label{fig:comparison}
\end{figure}

The Spearman rank correlation between model rankings on the full 208-item dataset and on the 94-item clean subset was $\rho = 0.532$, indicating that contamination substantially reshapes the apparent performance ordering; see Figure~\ref{fig:comparison}. To mention the most notable shifts: LLaMA~3.1~8B rose from rank~14 to rank~6, GPT-5.4~Mini dropped from rank~6 to rank~13, and Claude Opus dropped from rank~2 to rank~9. Another noteworthy finding was that the arithmetic-mean cloud--local gap collapsed from 35.8\% to 8.9\% on clean items (Table~\ref{tab:crowdsplit}).\footnote{To rule out the possibility that the gap collapse reflects a change in question-pool composition rather than the absence of contamination, we recomputed the cloud--local gap on the 48~items drawn directly from the original 208-item dataset that resolve after 1~September~2025 (i.e., excluding the 46~additionally collected items). On this nested subset---an estimate of the same quantity reported as 8.9\% in Table~\ref{tab:crowdsplit}, but based on only 48~items and hence noisier---the gap collapsed to 13.4\%, in line with the 8.9\% observed on all 94~clean items and confirming that the effect is attributable to contamination rather than to differences in question type or difficulty. Also, because training cutoff dates are approximate, we verified that the key contamination findings survive shifting all cutoff dates by $\pm 1$~month. Under both perturbations, the cloud--local gap remained large on the full dataset ($\approx 36\%$) and collapsed similarly on the nested clean items ($\approx 13\%$), mean accuracy differentials were stable (mean $\Delta\text{acc} \approx 0.17$), and Spearman rank correlations between full-dataset and out-of-window rankings remained around~0.5.}

\subsection{Symbolic regression}\label{sec:srres}

Before presenting the results, let us make explicit the argumentative work this analysis is meant to do. The near-equivalence of the logistic regression and the MLP (Sect.~\ref{sec:aggres}) shows that much of the aggregation benefit can be captured by a \emph{linear} combination of model outputs. But linearity alone does not show that the learned aggregator exploits the crowd \emph{as a crowd}: a linear aggregator could in principle concentrate all of its weight on the single best model, in which case what looks like aggregation would reduce to expert selection. Which models SR recruits, and how recruitment relates to individual performance, provides evidence on precisely this question: an aggregator performing expert selection should recruit models roughly in proportion to their individual performance.

That is not what we find. Across 50~independent SR runs, 11~models were selected in more than 50\% of runs (Table~\ref{tab:sr}). Selection frequency was only weakly correlated with individual Brier score rank ($r_s = 0.295$). The models selected in 100\% of runs included both the best-performing cloud models (Claude Sonnet, Claude Opus, Gemini Flash, Gemini Pro) and three of the worst-performing local models (Qwen~2.5~7B, LLaMA~3.1~8B, Phi-4~14B), producing a U-shaped selection pattern inconsistent with performance-weighted averaging: pairing the strongest cloud models with several of the weakest local ones is exactly what one expects if the aggregator is mining complementary error patterns, and exactly what one does not expect if it is picking experts. The lowest-complexity useful formula on the Pareto frontier was $\sigma(p_{\text{Gemini Pro}} - p_{\text{Claude Haiku}})$, a contrastive signal between two specific models---neither of which is the best individual model in the crowd.

\begin{table}[t!!]
\centering
\caption{Symbolic regression model selection frequencies across 50 independent runs (ranked by selection frequency). Brier rank refers to rank on the full 208-item dataset.}
\label{tab:sr}
\small
\begin{tabular}{l @{\hspace{15mm}} r @{\hspace{15mm}} r}
\toprule
Model & Selection (\%) & Brier rank \\
\midrule
Claude Sonnet~4.6   & 100 &  1 \\
Claude Opus~4.6     & 100 &  2 \\
Gemini Flash        & 100 &  3 \\
Gemini Pro          & 100 &  4 \\
GPT-5.4~Mini        & 100 &  6 \\
Qwen~2.5~7B         & 100 & 13 \\
LLaMA~3.1~8B        & 100 & 14 \\
Phi-4~14B           & 100 & 12 \\
Claude Haiku~4.5    &  98 &  5 \\
Gemini Flash-Lite   &  68 & 11 \\
GPT-5.2             &  54 &  7 \\
Mistral~7B          &  46 & 15 \\
DeepSeek-chat       &  28 &  9 \\
GPT-5.4             &  16 &  8 \\
Gemma~2~9B          &   0 & 10 \\
\bottomrule
\end{tabular}
\end{table}

To quantify how much of the learned aggregation benefit this minimal formula captures, we evaluated $\sigma(p_{\text{Gemini Pro}} - p_{\text{Claude Haiku}})$ directly as an aggregator, scoring it against the actual resolutions. On the 203~full-dataset items for which both constituent models returned a probability, it achieved BS = 0.231, against BS = 0.265 for the 15-model arithmetic mean on the same items---a reduction of 12.9\% (computed as $(0.265 - 0.231)/0.265$). On the 94-item clean subset, it achieved BS = 0.243. That a formula using only two models---one strong positive signal and one contrastive signal---can nearly replicate the performance of an aggregator learning from 15~models is a strong indication that the aggregation benefit comes from a small number of complementary error patterns rather than from complex nonlinear interactions among all 15 models.

\subsection{Logistic regression coefficient analysis}\label{sec:coef}

The symbolic regression results are suggestive but indirect---they show \emph{which} models are selected, not \emph{why} they got selected. For a more direct test, we inspected the logistic regression weights, which are interpretable by construction, and checked whether they reflect diversity or performance. For each model, we computed its mean pairwise Pearson correlation of squared errors with the other 14~models across the 94~clean items, as a measure of how redundant its error pattern is relative to the crowd. We then computed Spearman correlations between absolute logistic regression weight and (a)~diversity (negative mean error correlation, so that higher values mean more independent errors) and (b)~individual Brier score rank.

The results strongly favor the diversity interpretation: the Spearman correlation between absolute weight and diversity was $r_s = +0.482$, while the correlation with individual performance rank was $r_s = -0.075$, indicating that the aggregator learns to upweight models whose errors are independent from the crowd rather than models that are individually accurate. Notably, nine of the fifteen models receive negative weights, meaning the aggregator uses them \emph{contrastively}, in that when such a model assigns high probability, the aggregate is pulled downward. This is a further expression of error-pattern exploitation: a model's systematic biases can be informative even when its raw predictions are poor, provided those biases are sufficiently distinct from the rest of the crowd.

A terminological caution is in order concerning ``diversity.'' Throughout this paper, we use the term in the decorrelation sense just operationalized: a model contributes diversity to the extent that its errors are weakly (or negatively) correlated with those of the rest of the crowd. So understood, diversity is a separately measurable property of the crowd's joint error distribution, distinct from the spread-based notion of prediction diversity that figures in Page's Diversity Prediction Theorem \citep{Page2007}, and none of our claims rests on that theorem. As critics have pointed out, the theorem is an algebraic identity in which collective error, average individual error, and prediction diversity cannot be varied independently of one another, so it cannot by itself support the causal claim that making a crowd more diverse makes it more accurate \citep{ReiaFontanari2021,SiqueiraNetoFontanari2023}; indeed, \citet{ReiaFontanari2021} report forecasting data in which greater spread among the estimates goes hand in hand with \emph{larger} collective error. The relationship we report here is empirical, not derived from any identity: the learned weights track measured error decorrelation ($r_s = +0.482$) and not individual Brier performance ($r_s = -0.075$). It also parallels what has recently been observed in human crowds, where groups whose members produce less correlated probability judgments achieve higher accuracy even after controlling for the members' average individual accuracy \citep{DouvenKriegeskorteStinsonYing2026}.

\subsection{Comparison with human crowd}

The Manifold community probability provides two natural baselines. The \emph{final} market probability (at market close) reflects continuous updating right up to resolution; the \emph{cutoff-matched} market probability is the market price recorded at the time of each model's training cutoff, controlling for the informational advantage of more recent data.

On the full 208-item dataset, the final market probability achieved BS = 0.098, compared to 0.266 for the LLM arithmetic mean, which is a factor of 2.72.\footnote{The human comparison can use all 208~items because the benchmark is evaluated at the model's cutoff rather than at market close, thereby removing the market's general recency advantage. Any contamination of the LLM itself would tend to make the comparison more favorable to the LLM, not to the market. The exceptional fallback cases, for which cutoff-matching is imperfect, are addressed by the sensitivity analysis in footnote~\ref{fn:fallback}.} This large gap partly reflects the continuous-updating advantage: the final market price incorporates all information available before resolution, while LLMs reason from a frozen training snapshot.

\begin{table}[t!!]
\centering
\caption{LLM Brier scores versus Manifold market probability at each model's training cutoff date (208-item dataset). Both the LLM and the market-at-cutoff Brier scores are computed over each model's non-missing items (Sect.~\ref{sec:elicitation}), which is why the market column can differ slightly between models sharing a cutoff date. Final market BS = 0.098 (same for all models). Ratio = LLM BS / market-at-cutoff BS\@.}
\label{tab:cutoff}
\small
\begin{tabular}{l @{\hspace{11mm}} l @{\hspace{11mm}} r @{\hspace{11mm}} r @{\hspace{11mm}} l}
\toprule
Model & Cutoff & LLM BS & Market@cutoff & Ratio \\
\midrule
Claude Sonnet~4.6   & Aug 2025 & 0.273 & 0.132 & 2.06$\times$ \\
Claude Opus~4.6     & Aug 2025 & 0.284 & 0.133 & 2.14$\times$ \\
Claude Haiku~4.5    & Jul 2025 & 0.311 & 0.125 & 2.48$\times$ \\
Gemini Flash        & Jan 2025 & 0.286 & 0.142 & 2.01$\times$ \\
Gemini Pro          & Jan 2025 & 0.292 & 0.142 & 2.05$\times$ \\
Gemini Flash-Lite   & Jan 2025 & 0.377 & 0.143 & 2.64$\times$ \\
DeepSeek-chat       & Nov 2024 & 0.361 & 0.165 & 2.19$\times$ \\
GPT-5.4~Mini        & Sep 2024 & 0.328 & 0.204 & 1.61$\times$ \\
GPT-5.2             & Sep 2024 & 0.338 & 0.204 & 1.66$\times$ \\
GPT-5.4             & Sep 2024 & 0.347 & 0.204 & 1.71$\times$ \\
Qwen~2.5~7B         & Sep 2024 & 0.389 & 0.204 & 1.91$\times$ \\
Gemma~2~9B          & Jun 2024 & 0.362 & 0.222 & 1.63$\times$ \\
Phi-4~14B           & Jun 2024 & 0.381 & 0.221 & 1.73$\times$ \\
LLaMA~3.1~8B        & Dec 2023 & 0.394 & 0.232 & 1.70$\times$ \\
Mistral~7B          & Mar 2023 & 0.413 & 0.230 & 1.80$\times$ \\
\midrule
Arithmetic mean     & --- & 0.266 & 0.132\rlap{$^{\dagger}$} & 2.01$\times$ \\
\bottomrule
\multicolumn{5}{l}{\footnotesize $^{\dagger}$ Market at August 2025, the latest model cutoff in the ensemble.}
\end{tabular}
\end{table}

To isolate the contribution of this updating advantage, we fetched the historical market price at each model's training cutoff date from the Manifold API and computed the Brier score of each of those cutoff-matched probabilities. The results are shown in Table~\ref{tab:cutoff}. Even after this temporal matching, every individual LLM remained substantially less accurate than the market at the same point in time, with LLM Brier scores 1.61--2.64 times worse than the corresponding market probability (mean ratio: 1.95$\times$). For the LLM arithmetic mean compared to the market at the latest model cutoff in the ensemble (August 2025), the ratio was 2.01$\times$. This demonstrates that the human crowd advantage reflects genuine information aggregation---the ability of many active traders to integrate dispersed knowledge and update on each other's estimates---rather than merely having access to more recent data.

These results differ from \citet{Schoenegger2024}, who found their LLM crowd statistically indistinguishable from human forecasters. We come back to this in Section~\ref{human_diff}.

%------------------------------------------------------------
\section{Discussion}

\subsection{Main findings}

We found evidence for crowd wisdom effects in LLM aggregation: on the 94-item subset of questions resolving outside every model's training window, both the MLP and the logistic regression aggregator achieved lower Brier scores than all individual models and all classical aggregators. As in \citet{DouvenKriegeskorteStinson2026}, the mechanism appears to be the same as in human crowds \citep{LarrickSoll2006,DouvenKriegeskorteStinsonYing2026}: the aggregation benefit derives from exploiting error-pattern diversity rather than from weighting models by individual performance, as supported by our symbolic regression and coefficient analyses.

That said, LLM crowd wisdom is not yet at the level of human crowd wisdom, at least as measured against a continuously updated prediction market: even when the market probability is evaluated at the same point in time as each model's training cutoff---neutralizing the market's informational advantage---the market outperforms every LLM by a factor of 1.6--2.6. Section~\ref{human_diff} considers the source of this gap and explains why our result does not conflict with the findings of \citet{Schoenegger2024}.

Nevertheless, the gains from learned aggregation are both meaningful and statistically reliable, and they show how much additional benefit can be extracted, by comparatively simple learned aggregation, from the information LLMs carry. That a linear model performs comparably to the nonlinear model---and that the mechanism mirrors what has been found with human crowds---is the core finding of the paper. At the same time, so long as the constituent models reason from frozen training snapshots, without access to real-time information, there is a ceiling on how accurate any LLM crowd can be, regardless of how well it is aggregated.

\subsection{Crowd wisdom or expert selection?}\label{sec:woc_or_expert}

Taken together, the analyses of Sections~\ref{sec:srres} and~\ref{sec:coef} tell against a worry invited by the move from equal-weight pooling to learned aggregation: that the benefits we observe are not wisdom-of-crowds effects at all but the product of covert \emph{expert selection}, the learned aggregator in effect identifying the few best models in the ensemble and deferring to them, so that the exercise would belong to ensemble engineering rather than to social computing. Against this reading, the learned weights track error decorrelation rather than individual performance; negative weights allow individually weak models to contribute contrastively; and symbolic regression repeatedly recruits weak and strong models together. The remaining question is how such learned weighting should be situated within crowd-wisdom theory.

What the learned aggregators do, then, is what theorists of crowd wisdom have long said good aggregation should do: exploit the complementary structure of the members' errors \citep{LarrickSoll2006,DouvenKriegeskorteStinson2026}. It must be granted that learned weighting departs from the strictly democratic, equal-weight ideal of the \emph{Vox Populi}, and in that sense our aggregators occupy a middle ground between classical crowd wisdom and expert selection. But the departure runs in the direction of \emph{deeper} reliance on the crowd: the error-correlation structure that the weights track is information that exists only at the level of the ensemble and that no individual member possesses, whereas expert selection would dispense with the ensemble in favor of its best members. The parallel finding for human crowds---that groups whose members give less correlated judgments are more accurate, controlling for individual accuracy \citep{DouvenKriegeskorteStinsonYing2026}---suggests that this is a continuation of the collective-intelligence story rather than a break with it.

\subsection{Contamination and the cloud--local gap}

Training cutoff contamination was a pervasive and consequential confound. All models with sufficient within-cutoff items showed meaningfully higher accuracy on those items, with differentials up to 0.26. The consequences for model rankings were substantial ($\rho = 0.532$). Most strikingly, the cloud--local performance gap collapsed from 35.8\% to 8.9\% on clean items, suggesting that much of the apparent capability advantage of frontier models in probabilistic forecasting reflects knowledge of outcomes rather than superior probabilistic reasoning.

We observed a qualitative difference between model families in how contamination was expressed. Gemini models showed significantly elevated probability extremity on within-cutoff items (i.e., overconfident retrieval of known outcomes). Anthropic models showed comparable accuracy advantages without elevated extremity, suggesting that calibration training discouraged overconfident expression even when outcomes were known.

\subsection{Why does the human crowd outperform the LLM crowd?}\label{human_diff}

As noted, the cutoff-matched comparison places the LLM arithmetic mean at roughly 2 times the market's Brier score even
after controlling for the market's continuous-updating advantage. While, at first glance, this seems in tension with \citet{Schoenegger2024}, who found their LLM crowd statistically indistinguishable from a human crowd, the tension dissolves once the two baselines are compared directly.

Note that Schoenegger and colleagues compared against a \emph{forecasting tournament} aggregate: a set of human predictions made at a single point in time, by non-expert volunteers, on questions selected for the tournament. This is structurally very similar to our LLM elicitation: simultaneous, non-interactive, from generalist predictors with no special stake in accuracy. It is therefore not very surprising that LLMs can match this kind of human performance. Our baseline is fundamentally different: an actively maintained \emph{prediction market}, in which self-selected traders with domain knowledge bet repeatedly, react to each other's estimates and to new information, and collectively drive the market price toward a well-calibrated consensus. This is a much higher epistemic bar.

In other words, our findings and those of \citet{Schoenegger2024} are not in conflict but rather answer different questions: where Schoenegger and colleagues show that LLM crowds can match a panel of simultaneous human judgments, we show that they fall substantially short of a dynamic human market that aggregates dispersed knowledge through repeated interaction. The gap we observe in our study is not a failure of LLMs relative to humans in general, but a reflection of the difference between two very different human aggregation mechanisms. Whether LLM crowds could be brought closer to prediction market performance (perhaps through iterative deliberation protocols that allow models to observe and respond to each other's estimates) is an interesting question for future work.

\subsection{Limitations and future directions}\label{sec:limits}

The clean subset of 94~items is too small for stable ranking of individual models or for detailed analysis of subgroup differences. Future work should aim for at least 200--300~clean items, which would also allow training learned aggregators entirely within the clean data. Moreover, we used a single elicitation prompt without variation; different prompting strategies may affect both individual performance and inter-model error correlations. Also, we did not include very large open-weight models ($\geqslant\,$70B), which might behave differently from both the small local and frontier API models in our study.

Another limitation concerns the learned aggregators' training environment, flagged already in Section~\ref{sec:aggmeth}. The 160~non-clean items on which the MLP and logistic regression are trained are structured by training-cutoff contamination: models perform systematically better on items whose outcomes fall within their training window, producing a different pattern of errors than on genuinely novel questions. The learned weights are therefore fitted to a contaminated error landscape and then applied to a clean one, which is a form of distribution shift. That the aggregation advantage survives this shift (sign test: $62/94$, $p = .002$) is encouraging and suggests the learned strategy captures something general about the crowd's error structure. Nevertheless, whether the specific weights would remain stable across different training configurations (e.g., if trained on a fully uncontaminated dataset of comparable size) is an open question that
future work with larger clean datasets could address.

%==============================================================

\section*{Acknowledgments}
I am greatly indebted to two referees and an associate editor for exceptionally constructive feedback on a previous version.

\section*{Supplementary materials}
Supplementary Materials, containing the Julia \citep{Bezanson2017} code and the data used for the study, can be downloaded from this repository:  \url{https://osf.io/8dng6/overview?view_only=3c535d46fe924def902e5579983313cc}.

%==============================================================

\bibliographystyle{ACM-Reference-Format}
\bibliography{references}

\end{document}